%% file: main.tex
\documentclass[pmlr]{jmlr}

\usepackage{longtable}

\usepackage{booktabs}
\usepackage[load-configurations=version-1]{siunitx} 

 \usepackage{amsmath}
 \usepackage{amssymb}
 
 \usepackage{times}  
 \usepackage{helvet}  
 \usepackage{courier}  
 \usepackage{url}  
 \usepackage{graphicx}  
 \usepackage{color}
 \usepackage{amsfonts}
 \usepackage{amsmath}
 \usepackage{dsfont}
 \usepackage{amssymb} 

 \DeclareMathOperator*{\argmax}{arg\,max}
 
 \newcommand{\paratitle}[1]{\vspace{1.5ex}\noindent \textbf{#1}}

\theorembodyfont{\upshape}
\theoremheaderfont{\scshape}
\theorempostheader{:}
\theoremsep{\newline}

\title[Blood Glucose Prediction]{Using Contextual Information to Improve Blood Glucose Prediction}

\author{\Name{Mohammad Akbari} \Email{akbari@nyu.edu} 
      New York University\\
      \AND
      \Name{Rumi Chunara} \Email{rumi.chunara@nyu.edu} 
      New York University\\
   } 

\begin{document}

\maketitle

\begin{abstract}
Blood glucose value prediction is an important task in diabetes management. While it is reported that glucose concentration is sensitive to social context such as mood, physical activity, stress, diet, alongside the influence of diabetes pathologies, we need more research on data and methodologies to incorporate and evaluate signals about such temporal context into prediction models. Person-generated data sources, such as actively contributed surveys as well as passively mined data from social media offer opportunity to capture such context, however the self-reported nature and sparsity of such data mean that such data are noisier and less specific than physiological measures such as blood glucose values themselves. Therefore, here we propose a Gaussian Process model to both address these data challenges and combine blood glucose and latent feature representations of contextual data for a novel multi-signal blood glucose prediction task. We find this approach outperforms common methods for multi-variate data, as well as using the blood glucose values in isolation. Given a robust evaluation across two blood glucose datasets with different forms of contextual information, we conclude that multi-signal Gaussian Processes can improve blood glucose prediction by using contextual information and may provide a significant shift in blood glucose prediction research and practice.

\end{abstract}

\input{introduction.tex}

\input{method.tex}
\input{experiments.tex}

\input{appendix.tex}

\small
\bibliography{aaai_ref}
\end{document}

%% file: introduction.tex
\section{Introduction}
Blood glucose prediction is an important problem. It is necessary in order to anticipate and mitigate extremely high (hyperglycemic) or low (hypoglycemic) glucose events. The introduction of sensors such as Continuous Glucose Monitors (CGMs), have allowed for data-driven, short-term glucose prediction methods (for example, \cite{plis2014machine,fox2018deep}). Beyond improvement of such time-series modeling methods, recent review of blood glucose prediction work has indicated major areas where blood glucose prediction can be improved. This has included accuracy of the sensors, delay of insulin action and glucose level estimation by the continuous glucose monitoring (CGM) system, and the lack of models that account for social, contextual and emotional factors that are known to affect glucose concentration (\cite{oviedo2017review}).

The mechanism of action for these multiple contexts that can have a prominent impact on blood glucose levels, such as  activities, mood, emotional states, lifestyle, etc., can vary. For example mood and depression can act on blood glucose control via metabolic control, poor adherence to medication and diet regimens, changes in quality of life and/or healthcare expenditures (\cite{lustman2005depression}). Physical exercise can interact directly with glucose metabolism improves glucose metabolism in lifestyle-related diseases like type 1 diabetes (\cite{sato2003physical}). At the same time, there exists limited research in incorporating and examining multiple types of contextual information into a blood glucose prediction task, in part due to lack of data on these contexts along with blood glucose data.

Today, capturing this contextual information is becoming possible, through the increasing types of sensors via which many types of explicit and implicit information are shared directly by individuals. Some studies have included additional input signals related to physical activity, for example through commercial sensors to infer energy expenditure via acquiring data on skin temperature, heat flux, and galvanic skin response (\cite{mignault2005evaluation,oviedo2017review}). However, energy expenditure is highly dependent on the type of physical activity, and as well, accurate determination of energy expenditure is still a major challenge. While several simulation-based/physiological modeling approaches for blood glucose prediction have incorporated exercise models and gastric dynamics,
data-driven models that incorporate other additional signals need more exploration. In sum, studies that include additional signals are low in number, and they do not cover
the full range of possible context/side-information that have been demonstrated
to be related to blood glucose levels, such as emotional state.

Person-Generated Health Data (PGHD, which includes social media, wearable sensors, self-reported information in forms of surveys, etc.) is increasingly available and can potentially illuminate some of these factors that have been under-represented in empirical blood glucose prediction. However, modeling such data is a non-trivial task due to several factors including: 1) \textbf{Data quality.} PGHD is known as a highly varied, noisy, and sparse information source (\cite{agarwal2011sentiment}). Moreover, missing data is an intrinsic aspect of PGHD since people may not persistently or regularly report their health data online or record them in mobile applications. While there are several motivations for users to record and report their health data, in most cases, people are not sufficiently keen to generate data regularly, or they may self-censor the content due to privacy concerns (\cite{de2014seeking,huang2017high}), resulting in very sparse observations. Thus, the data is not comparable to controlled clinical data or continuous body sensors (such as continuous glucose monitoring), and missing data for extended and irregular periods of time is a major challenge. Furthermore, the self-reported nature of PGHD means the data is often unstructured. Specifically for text-based data, this is challenging as the latent meaning must be understood withstanding short test, slang and colloquial phrases. 2) \textbf{Temporal correlations.} Physiological attributes smoothly evolve over time. The temporal progression of such attributes suggests that these values gradually changes over time. However, this evolution can involve several periodic patterns of short- and long-term dependencies. Moreover, PGHD is generated irregularly, as described above. Therefore, how to model these temporal dependencies poses another challenge. 3) \textbf{Multi-signal temporal data.} Health and lifestyle information from PGHD can be in the form of both multi-view as well as multi-modal information, over time. For example some contextual features, such as insulin dose occurence, may provide multiple views into blood glucose (just like image-text pairs which have the same semantic meaning). Other contextual data may be multi-modal and/or multi-variate (providing different covariates such as mood, that relate to blood glucose, and are provided in different modalities such as text sentiment). Hence, developing a model that is capable of capturing and fusing multiple signals that encompass both multiple views and modalities from sparse, noisy, and multi-variate temporal data is a specific challenge.

To address all of these technical challenges and fill the need for blood glucose prediction with contextual information, here we implement a Gaussian Process model for this novel \textit{multi-signal} task, which accounts for multiple contexts, mechanisms, and irregular context data from PGHD. We test our method using two types of PGHD data which also represent multiple types of context; actively contributed surveys alongside CGM data, as well as social media data posts of glucose and associated context. Social media data is less specific, and while the amount of data may be very limited if specific features were curated (e.g. only physical activity or diet mentions), we show that a user’s continuous context, defined by latent features from their stream, can still augment blood glucose prediction. Surveys are actively contributed, so although they provide more specific information, they may suffer from recall biases.

\paragraph{Clinical Relevance} We evaluate how contextual information can be used in a blood glucose level prediction. Though several years of medical research have shown the importance of daily lifestyle and context on blood glucose variation, in addition to diabetes pathologies (\cite{oviedo2017review,sato2003physical}), this premise has not been investigated in detail due to data and methodological limitations. We evaluate this idea using two different types of data, each providing different forms of contextual information and content. Extensive experiments demonstrates the effectiveness of the proposed approach in predicting future blood glucose values. While we demonstrate this approach on available datasets, this work may provide a shift in thinking for glucose prediction research and practice, potentially informing development of new tools for capturing more specific context information alongside physiologic data.

\paragraph{Technical Significance}
In order to solve this problem, to incorporate and evaluate contextual information in blood glucose value prediction, this work demonstrates technical innovation that differs from prior blood glucose research. The proposed non-parametric, non-linear approach addresses the problem of fusing multiple signals, (i.e. blood glucose time series data and contextual information) through combining information in consistent latent space representations. We note that this is different from multi-view or multi-modal tasks, as the contextual information is different in data mode than glucose values, provides added information via relevant covariates as well as 
other views into blood glucose values. We demonstrate that this multi-signal Gaussian Process approach is superior to other common approaches for combining multivariate data, and to just using the blood glucose data on it's own, for blood glucose prediction. Finally, this allows us to evaluate and compare different types of contextual information, which can inform future detailed studies of causal mechanisms.

%% file: method.tex
\section{Problem Formulation}

The problem we study in this paper is to estimate the next blood glucose level of patients based on their historical data. 
Intuitively, blood glucose prediction is modeled as a regression task, where we forecast the future value of blood glucose based on a window of historical data. This prediction is then computed recursively to estimate all the values in the prediction window. Besides, in blood glucose prediction, the management and prevention of hypo/hyperglycemic events is particularly of interest (instead of predicting exact blood glucose values) (\cite{sparacino2007glucose}), therefore we also aim at predicting blood glucose level in hypoglycemic (low), euloglycemic (normal), and hyperglycemic (high) glucose range categories.
Following, we first introduce the notation used in this paper and then formally define the problem of blood glucose level prediction.  

Let $\mathcal{U} = \{ u_1,u_2,\ldots,u_n \}$ denote a set of $n$ different users. For each user $u$, the historical blood glucose values are given as $\mathbf{v}^u = [ v^u_1, v^u_2, \ldots, v^u_t ]$, where $v^u_k$ denotes the blood glucose value of the user $u$ at timestamp $k$. $v^u_k$ ia a positive value if the $k$-th blood glucose value is observed, otherwise $0$ (a missing value). We also have a target series $\mathcal{Y}^u$ of the same length where $\mathbf{y}^u = [ y^u_1, y^u_2, \ldots, y^u_t ]$ shows the blood glucose level of the user $u$ (blood glucose value for  regression task or blood glucose level in three categories for classification task) at the same observation points. Given the social/contextual side information of a user denoted by $\mathbf{S}^u \in \mathbb{R}^{t \times s}$, where $t$ and $s$ denote the total number of observation points and the number of features from side information, respectively, we aim to learn a non-linear mapping to predict the next blood glucose level in the target series, namely $y^{\ast}_{t+1}$. 

Based of the above discussion, we now formally define the problem of \textbf{blood glucose level prediction with complementary side information} as, 

\textit{Given a set of user $\{ \mathcal{U}^i \}_{i=1}^{n}$, their historical blood glucose values $\mathbf{v}^u$, their target series $\mathbf{y}^u$ and side information $\mathbf{S}^u$,  we aim at learning a parametric probability measure:
\begin{align}
y^{\ast}_{t+1} = \argmax_{y_{t+1} \in \mathcal{L}} p(y_{t+1} \vert \mathbf{v}^u, \mathbf{y}^u, \mathbf{S}^u, \boldsymbol\theta)
\end{align}
where $\boldsymbol\theta$  denotes a vector of all parameters of the model to be learned.}

\section{Methodology}
In this section, we explain our method to model and make predictions from time series of user blood glucose values. In particular, we exploit Gaussian Processes, a non-parametric and non-linear method, to better capture temporal patterns of blood glucose values of users. Gaussian processes have recently been applied to time series analysis due to their flexibility and ability to deal with missing data naturally (\cite{futoma2017learning,chung2018mixed}). In fact, Gaussian Processes have been used in blood glucose prediction studies (\cite{valletta2009gaussian}, \cite{albers2017personalized}). The work in these papers are different than the goal here though, based on combining contextual information by concatenating it with the blood glucose data, and considers only select context variables that are directly related to carbohydrate consumption and physical activity. 

Here, as we consider the problem of how to broadly combine context data from PGHD and blood glucose data, it should be noted that this does not directly map to the multi-view or multi-modal paradigm; there are specific limitations and differences of existing methods in relation to the task at hand. Many researchers have looked into the general problem of combining information from multiple sources, especially in multimedia information retrieval. Two major approaches are early fusion (\cite{snoek2005early,xu2011efficient}) and late fusion (\cite{ye2012robust}). Early fusion methods, such as the work in \cite{snoek2005early}, construct a joint feature space by merging all the extracted features from different views into a joint feature vector by concatenating all features. These approaches often are limited in noisy and sparse scenarios, as noisy or missing features extracted from one view will affect the joint features and result in reducing performance of the prediction model. Besides, the multiple types of features considered can have different semantic interpretation, like our case, of one feature being time series of blood glucose measures and a second contextual features coming from social media text; simply merging all feature views would bring in a certain extent of noise and ambiguity. In late fusion (\cite{snoek2005early,ye2012robust}), on the other hand, a model is learned based on each view separately and then results are integrated to make the final decision. For example, \cite{kapoor2005multimodal} proposed a unified approach, based on a mixture of experts, for classifying interest of children in a learning scenario using multiple modalities. This approach generates separate class labels corresponding to each individual modality. The final classification is based upon a hidden random variable, which probabilistically combines the sensors. Accuracy of the late fusion approach can suffer because an individual feature space might not be sufficiently descriptive to represent the complex semantics of a task, especially in noisy, sparse and missing data scenarios. Therefore, separate results would be suboptimal and the integration may not result in the desired outcome. Finally, while it has been previously recognized that social media data can be sourced for multi-view features (\cite{tang2013unsupervised}), these have not been utilized for temporal applications. Bringing together multiple signals in one latent space through Gaussian processes specifically addresses challenges of the combining context which presents noisy and sparse, and temporally inconsistent information, with multiple views and multiple modes, with blood glucose.

\subsection{Preliminaries}
\label{sec::preliminaries}
Gaussian processes are the state-of-the-art, non-parametric regression method. A Gaussian Process is formally defined as a collection of random variables where any subset of them taken together jointly form a (multivariate) Gaussian distribution. A useful intuition is to view a Gaussian Process as a distribution over functions (function-space view \cite{rasmussen2004gaussian}). Since a Gaussian Process is a distribution over functions, sampling from a Gaussian Process results in the draw of a single function. 
Gaussian Processes follow the Bayesian paradigm of updating prior beliefs based on observed data to form posterior distributions. In the case of Gaussian Processes, these prior and posterior distributions are distributions over functions, and therefore the Bayesian inference that takes place occurs in function space.

Let $\mathcal{D} = \{ (\mathbf{x}_1,y_1),(\mathbf{x}_2,y_2),...,(\mathbf{x}_n,y_n)\}$ denote a dataset of input values (i.e., feature vectors) and their corresponding outcomes. A Gaussian process is defined as a stochastic model over the latent function $f$ that maps the inputs in $\mathbf{x}$ to their corresponding response variables. Formally, a Gaussian process can be stated as, 
\begin{align}
f(\mathbf{x}) \approx \mathcal{GP} ( m(\mathbf{x}), k(\mathbf{x}_i,\mathbf{x}_j) ), 
\end{align}
where $m(\mathbf{x})$ is the mean function and $k(\mathbf{x}_i,\mathbf{x}_j)$ is the kernel or covariance function, which describes the degree to which the output values of $f$ covariate at locations $\mathbf{x}_i$ and $\mathbf{x}_j$. 
Without loss of generality we can assume a zero prior mean function everywhere, i.e. $m(\mathbf{x}) = 0$. Hence, the kernel specifies the high-level assumptions of the underlying function $f$.

Following Bayes' theorem, to make a prediction on new data $\mathbf{x}^{\ast}$ the predictive posterior can be computed by, 
\begin{align}
p(f^{\ast} \vert \mathbf{X}, \mathbf{x^{\ast}}, \mathbf{y}) = \int_f p(f^{\ast} \vert \mathbf{x^{\ast}}, \mathbf{f}) p( \mathbf{f} \vert \mathbf{X}, \mathbf{y}) df, 
\end{align}
where  $ p( \mathbf{f} \vert \mathbf{X}, \mathbf{y}) = \frac{p(\mathbf{y} \vert \mathbf{f}) p(\mathbf{f})}{p(\mathbf{y} \vert \mathbf{X})}$ is the posterior over the latent space. 
The likelihood of the data $\mathbf{X}$ given the latent space $f$ is,
\begin{align}
p(\mathbf{X} \vert \mathbf{f}, \theta) = \frac{1}{A} \exp{(-\frac{1}{2} tr(\mathbf{K}^{-1} \mathbf{X}\mathbf{X}^T))}
\end{align} 
where $A =\sqrt{(2 \pi)^{ND_x} \times \vert \mathbf{K} \vert^{D_x}}$ is the normalization factor and $\mathbf{K} \in \mathbb{R}^{n \times n}$ is the kernel matrix defined on $\mathbf{X}$, i.e., $\mathbf{K}_{ij} = k(\mathbf{x}_i, \mathbf{x}_j )$. 


The key advantage of this approach is the use of Bayesian learning for mitigating problems arising from over-fitting with small data as well as avoiding the need to select parameters of the function approximators.
In the case of binary classification, the output of latent function $f(x)$ is then squashed into the range $[0,1]$ via a logistic function $\pi(\mathbf{x}) \doteq p(y = 1|\mathbf{x}) = \sigma(f(\mathbf{x}))$ in a similar way to logistic regression classification.

\subsection{Multi-Signal Gaussian Process with Side Information}

In this section we present how to incorporate complementary information sources to Gaussian processes and show the corresponding inference and optimization in the proposed model. We limit our discussion to the case that we have just one complementary signal. Extension to multiple signals can be easily inferred.

In the case that our input comprises from two input sequences, i.e., $\mathbf{x}_i = (\mathbf{v}_i, \mathbf{s}_i)$ , we first learn a unified latent space $\mathbf{Q}$ which preserves an underlying pattern existing in both views. We then build a classifier on the top of the learnt space. We use Gaussian processes to parameterize two functions $f_v: \mathbf{Q} \rightarrow \mathbf{V}$ and $f_s: \mathbf{Q} \rightarrow \mathbf{S}$ which map the shared latent space to two input signals. Here $\mathbf{V}$ denotes the input space representing blood glucose associated values and $\mathbf{S}$ represents social/contextual information of users. We base our model on a conditional independence assumption that, given the latent variable $\mathbf{Q}$, the two inputs $\mathbf{V}$ and $\mathbf{S}$ are independent.  Thus, based on discussion in the previous section \ref{sec::preliminaries}, the predictive posterior can be written as,
\begin{align}
\label{eq::predictive-posterior}
P(\mathbf{Q}, \theta_v, \theta_s \vert \mathbf{V}, \mathbf{S}) \propto p(\mathbf{V} \vert \mathbf{Q}, \theta_v) p(\mathbf{S} \vert \mathbf{Q}, \theta_s) p(\mathbf{Q}),
\end{align}
where the first and second terms are likelihood of data given the latent space and the third term is the prior on the latent vectors which can be modeled via a Gaussian prior as, 
\begin{align}
p(\mathbf{Q}) = \frac{1}{\sqrt{2 \pi}} \exp{( - \frac{1}{2} \sum_i \Vert \mathbf{q}_i \Vert^2)}.
\end{align}

Learning model parameters is often performed by minimizing the negative log posterior, i.e., Eq.(\ref{eq::predictive-posterior}) with respect to latent space, i.e., $\mathbf{Q}$, and the hyper-parameters, i.e., $\theta_v, \theta_s$, which is, 

\begin{align}
\label{eq::negative-log-posterior}
\mathcal{L} &= \mathcal{L}_v + \mathcal{L}_s + \frac{1}{2} \sum_i \Vert \mathbf{q}_i \Vert^2, \\
\mathcal{L}_v &= \frac{D_v}{2} \ln{ \vert \mathbf{K}_v \vert } + \frac{1}{2} tr(\mathbf{K}_v^{-1} \mathbf{V}\mathbf{V}^T), \\
\mathcal{L}_s &= \frac{D_s}{2} \ln{\vert \mathbf{K}_s \vert }+ \frac{1}{2} tr(\mathbf{K}_s^{-1} \mathbf{S}\mathbf{S}^T)).
\end{align}

Minimizing Eq.(~\ref{eq::negative-log-posterior}) with respect to $\mathbf{Q}$, $\theta_v$, and $\theta_s$ will result in a low dimensional space which is shared by both $\mathbf{V}$ and $\mathbf{S}$ inputs. 
Here we utilize an exponential (RBF) kernel to define the similarity between two data points $\mathbf{x}_i$, $\mathbf{x}_j$ as, 
\begin{align}
    k(\mathbf{x}_i, \mathbf{x}_j) = \theta_1 exp(-\frac{\theta_2}{2} \Vert \mathbf{x_i} - \mathbf{x_j}\Vert^2)
\end{align}

\subsection{Optimization}
We employed the Scaled Conjugate Gradient (SCG) to learn the optimal latent representation $\mathbf{Q}$. Specifically, the gradients of $\mathcal{L}_v$ and $\mathcal{L}_s$ can be computed as, 
\begin{align}
\label{eq::gradient-LV}
    \frac{\partial \mathcal{L}_v}{\partial \mathbf{q}_i} &= \frac{1}{2} \Big( n \mathbf{K}_v^{-1} - \big( \mathbf{K}_v^{-1} \mathbf{V}\mathbf{V}^T\mathbf{K}^{-1}_v \big) \Big)\frac{\partial \mathbf{K}_v}{\partial \mathbf{q}_i} \\
    \label{eq::gradient-LS}
    \frac{\partial \mathcal{L}_s}{\partial \mathbf{q}_i} &= \frac{1}{2} \Big( n \mathbf{K}_s^{-1} - \big( \mathbf{K}_s^{-1} \mathbf{S}\mathbf{S}^T\mathbf{K}^{-1}_s \big) \Big)\frac{\partial \mathbf{K}_s}{\partial \mathbf{q}_i}. 
\end{align}

The gradient of Eq.(\ref{eq::negative-log-posterior}) with respect to the latent representation is the sum of Eq.(\ref{eq::gradient-LV}) and Eq.(\ref{eq::gradient-LS}) plus the gradient of the third part in Eq.(\ref{eq::negative-log-posterior}).

%% file: experiments.tex
\section{Experiments}
In this section, we present the experimental details to verify the effectiveness of our proposed framework. We implement the proposed approach on two different types of data, each providing different types of contextual information: 1) high-resolution glucose sensor data with linked survey information on diet and physical activity context information (``CGM data'') and 2) social media data, in which users provide passive data that we use as context. The high-resolution CGM data along with its surveys is more structured and continuous, yet only available for a very small population sample. The social media data is available from a larger and broader set of users, and has passively contributed context information which decreases recall and information biases. However, this data is much more sparse, so we also evaluate the results according to different levels of sparsity. It should also be noted that the CGM data is from users with type 1 diabetes only, while we do not precisely know what kind of diabetes the social media users have. However, given the close attention to insulin management and social features that span medication and other factors that are relevant to type 1 diabetes management (see previous exploration of this data (\cite{akbari2018user}), we deem it fair to consider both of these datasets in this task.

In the following, we first benchmark our method on sensor data collected by CGM devices. For the CGM data we focus on two tasks of i) predicting blood glucose values for next $30$ minutes (regression task), and ii) predicting glycemic events in the same window (classification task). Due to sparsity, we only focus on predicting the onset of glycemic events in the social media data. However, we also conduct experiments to measure the impact of sparsity on prediction performance in this dataset. For both datasets, we first explain different features extracted from contextual data and then compare our method against state-of-the-art baselines. 


\subsection{Clinical Data: CGM Data}
To evaluate our framework on a continuous glucose dataset which also has some contextual information, We utilized the OhioT1DM dataset (\cite{marling2018ohiot1dm}) which includes data collected from six patients with type $1$ diabetes. For each subject, the following data were collected: a blood glucose level from continuous glucose monitor (CGM) every $5$ minutes; periodic finger sticks blood glucose levels from; insulin doses, both bolus and basal; self-reported meal times with carbohydrate estimates; self-reported times of sleep, work, and exercise; and 5-minute aggregations of heart rate, galvanic skin response (GSR), skin temperature, air temperature, and step count. To forecast blood glucose values, we used the historical values of blood glucose collected via CGM as main time series signal and insulin doses, both bolus and basal and self-reported meal times with carbohydrate estimates as contextual side information. Due to misalignment of values in time dimensions, we do not use the other side information sources (these can all be binned to larger time steps in future studies; we tried to use exact times where possible for this initial study). 

\subsubsection{Predictive Accuracy}
Here we describe the baselines used for evaluating our method on the regression and classification tasks.
For the classification task, we compared the following different approaches:
\begin{description}
  \item[$\bullet$ LR:] As a first baseline we used a logistic regression classifier, which is a linear classifier~\cite{shumway2011time}. Contextual features are included in the input vector.
  \item[$\bullet$ KCCA:] A kernelized version of Canonical Correlation Analysis (CCA), and thus this model can conduct non-linear dimensionality reduction for data of two views (\cite{thompson2005canonical,akaho2006kernel}).
  Canonical correlation is the most appropriate baseline to use, as it is applied to extract common features from a pair of multivariate data. The kernalized version also is selected to allow for non-linear relationships and to be more comparable to the proposed approach.
  \item[$\bullet$ GP:] This is the simplest model based on Gaussian processes and models blood glucose level detection as a time series prediction task using (only) historical blood glucose values.
  \item[$\bullet$ GP+Context:] Our proposed approach that employs context - learned in a unified latent space with blood glucose values. 
\end{description}

Table~\ref{tbl::performancre-result-cgm-classification} and Figure~\ref{fig::performancre-result-cgm-regression} (Appendix A) show the results of different methods on regression and classification tasks, respectively. In the regression task, GP+Context shows the lowest error. For the classification task, it is first noted that precision for detection of hypoglycemic events is the lowest. This could be due to the relative rarity of such events in the dataset  (\cite{marling2018ohiot1dm}). As show in Table~\ref{tbl::performancre-result-cgm-classification}, the performance of GP+Context approach in detecting abnormal glycemic events, i.e., hypoglycemic and hyperglycemic, is superior to the other baselines. This is important as the proposed method can detect adverse levels. KCCA demonstrates higher overall performance as compared to other baselines while it also achieves highest performance in the euglycemic case. In the regression task (forecasting blood glucose values instead of glycemic ranges), KCCA and LR achieve the worst results. This can be explained by the fact that they cannot model the temporal dependencies among blood glucose values. GP+Context outperforms other baselines which is attributed to the fact that it models temporal dependencies between blood glucose values as well as utilizes the available contextual information to better predict future blood glucose values. Overall, these observations demonstrate that the proposed model can learn an effective latent space from both time series data and contextual information.


\begin{table*}[h]
	\small
	\centering
	\caption{Comparison of different approaches on glycemic event detection on CGM dataset}
	\label{tbl::performancre-result-cgm-classification}
	\begin{tabular}{c|c|c|c|c|c|c|c|c|}
		\cline{2-9}
		& \multicolumn{2}{c|}{\textbf{Hypoglycemic}} & \multicolumn{2}{c|}{\textbf{Euglycemic}} & \multicolumn{2}{c|}{\textbf{Hyperglycemic}} & \multicolumn{2}{c|}{\textbf{Overall}} \\ \cline{2-9} 
		& Precision         & Recall         & Precision         & Recall         & Precision          & Recall         & Precision           & Recall          \\  \hline
		\multicolumn{1}{|l|}{\textbf{LR}}   & $0.18$ & $0.76$ & $0.53$ & $0.54$ &  $0.73$ & $0.68$  & $0.51$ & $0.63$  \\ \hline
		\multicolumn{1}{|l|}{\textbf{KCCA}}   & $0.26$ & $0.82$ & $\mathbf{0.86}$ & $\mathbf{0.75}$ &  $0.74$ & $0.75$  & $0.72$ & $0.76$  \\ \hline
		\multicolumn{1}{|l|}{\textbf{GP}}   & $0.12$ & $0.67$ & $0.82$ & $0.63$ &  $0.70$ & $\mathbf{0.82}$  & $0.58$ & $0.74$  \\ \hline
		\multicolumn{1}{|l|}{\textbf{GP+Context}} & $\mathbf{0.28}$ & $\mathbf{0.86}$ & $0.83$ & $0.70$ &  $\mathbf{0.79}$ & $0.77$  & $\mathbf{0.79}$ & $\mathbf{0.77}$  \\ \hline
	\end{tabular}
\end{table*}

\subsection{Social Media Data}

\subsubsection{Data Collection}

\begin{figure}
\centering
\includegraphics[width=0.6\columnwidth]{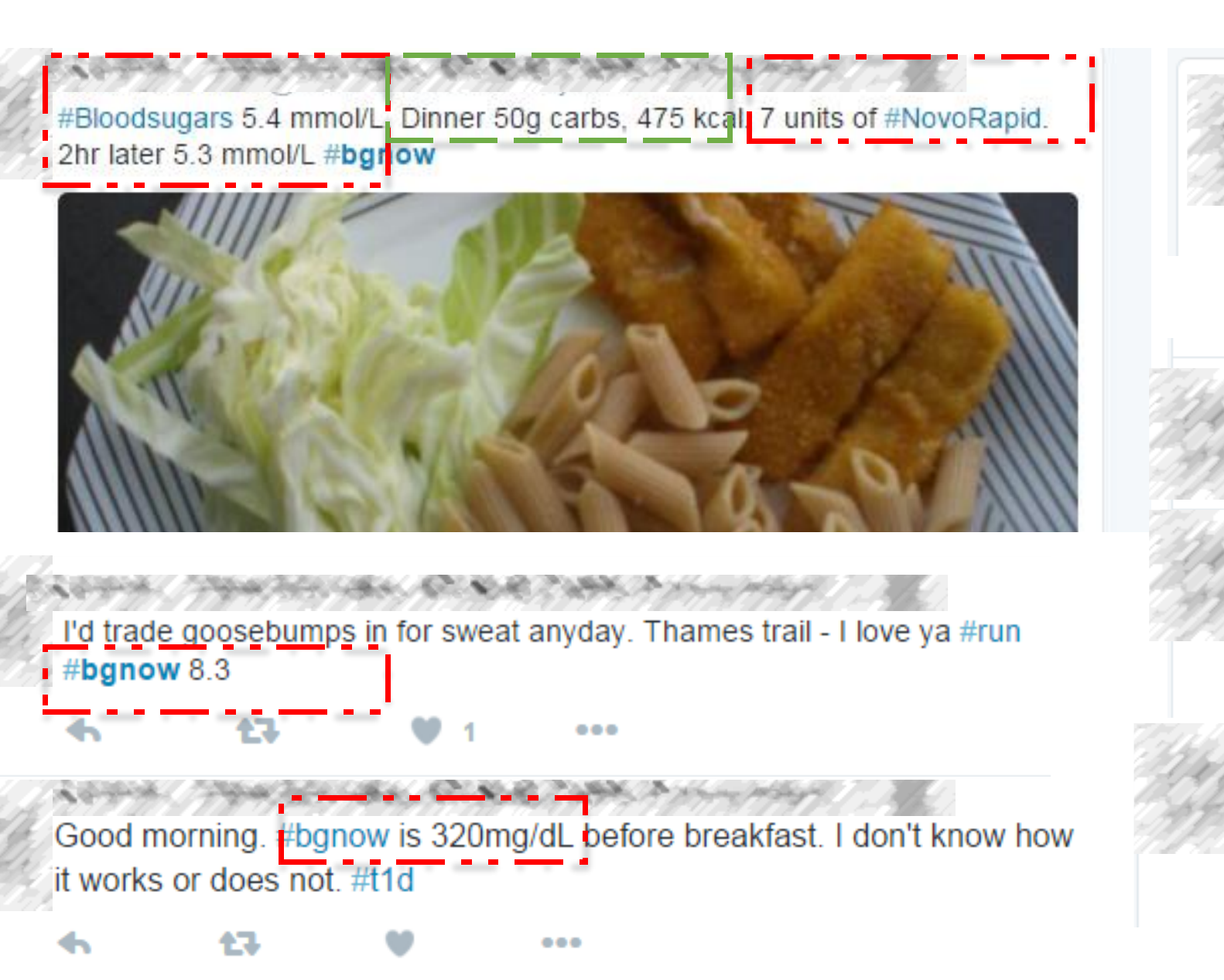}
\caption{Example tweets from the \#bgnow hashtag; the identity of users have been removed due to privacy concerns.}
\label{fig::bgnow-tweets-examples}
\end{figure}

Another available dataset which provides blood glucose values in concert with contextual information comes from social media. We collected a dataset using the hashtag `\#bgnow', which is widely-used by diabetics to share their blood glucose (BG) levels. Figure~\ref{fig::bgnow-tweets-examples} shows examples of tweets with this hashtag. As can be seen, individuals utilize this hashtag for self-reporting their BG values on Twitter. As such, we consider these values as a sparse BG time series. For each user, we construct his/her BG trajectories based on his/her self-reported BG values. To construct the dataset, we initially collected all tweets containing `\#bgnow'. 

\begin{table*}[h]
\small
\centering
\caption{Dataset Statistics}
\begin{tabular}{|c|c|c|c|c|c|}
\hline
\# Users & \# tweets &  \# BG values & \# hypogelycemic & \# euloglycemic & \# hyperglycemic \\ \hline
   $306$ & $1,909,171$ & $12,014$ & $1,936$ & $5,549$ & $4,529$ \\ \hline
\end{tabular}
\label{tbl::datsset-statistics}
\end{table*}

We next used these to find users with recent timeline data; we identified all distinct users who posted via this hashtag, and then collected twitter timelines of these users for a one year period\footnote{This is due to limitation of Twitter API which only retrieves latest $3,200$ tweets for each user.}. This allows us to analyze their online activities on Twitter alongside their blood glucose values. To remove the effect of non-active users, we removed users who posted less that $5$ blood glucose values as they do not have sufficient historical information. This results into a dataset of $306$ individuals and a total of $12,014$ blood glucose values posted by them. These users have posted about $39$ number of blood glucose values, in average, where the average days between two posts is equal to $11$ days. 


To create the ground-truth, we next extracted the reported blood glucose values from the included tweets. To do so, we defined a set of lexical rules to check the neighborhood window around the hashtag `\#bgnow' and looks for numerical values and measurement units such as `320mg/dl'~ (see Figure~\ref{fig::bgnow-tweets-examples}). These values next were mapped to their corresponding categories of: hypoglycemic (less than $70$ mg/dl), euloglycemic ($70-180$ mg/dl), and hyperglycemic (more than $180$ mg/dl) ranges. Table~\ref{tbl::datsset-statistics} shows the statistics of our dataset. 

Further, the text-content of tweets were utilized to extract features which we consider as side information from social media. The contextual information available in social media platforms are available in form of unstructured text, i.e., users post on the network. We extract social context of users from these posts. To represent user context we extracted two types of features: user-centric and content-centric features (see Appendix B). 

\begin{table*}[h]
\small
\centering
\caption{Comparison of different approaches on BG level prediction on Social Media Data.}
\label{tbl::performance-results}
\begin{tabular}{c|c|c|c|c|c|c|c|c|}
\cline{2-9}
                           & \multicolumn{2}{c|}{\textbf{Hypoglycemic}} & \multicolumn{2}{c|}{\textbf{Euglycemic}} & \multicolumn{2}{c|}{\textbf{Hyperglycemic}} & \multicolumn{2}{c|}{\textbf{Overall}} \\ \cline{2-9} 
                           &Precision         &Recall         &Precision         &Recall         &Precision     &Recall         &Precision         &Recall          \\  \hline
\multicolumn{1}{|l|}{\textbf{LR}}   & $0.38$ & $0.06$ & $0.54$ & $0.76$ &  $0.51$ & $0.45$  & $0.50$ & $0.52$  \\ \hline
\multicolumn{1}{|l|}{\textbf{KCCA}}   & $0.53$ & $0.10$ & $0.54$ & $0.73$ &  $0.51$ & $0.48$  & $0.53$ & $0.53$  \\ \hline
\multicolumn{1}{|l|}{\textbf{GP}}   & $0.39$ & $0.49$ & $0.41$ & $0.36$ &  $0.40$ & $0.33$  & $0.40$ & $0.40$  \\ \hline
\multicolumn{1}{|l|}{\textbf{GP+Social}} & $0.71$ & $0.43$ & $0.62$ & $\mathbf{0.84}$ & $0.63$ & $0.51$  & $0.64$ & $0.63$  \\ \hline
\multicolumn{1}{|l|}{\textbf{GP+Context}} & $\mathbf{0.86}$ & $\mathbf{0.53}$ & $\mathbf{0.88}$ & $0.83$ &  $\mathbf{0.72}$ & $\mathbf{0.88}$ & $\mathbf{0.82}$ & $\mathbf{0.80}$  \\ \hline
\end{tabular}
\end{table*}

\subsubsection{Predictive Accuracy}
We utilized the same set of baselines to benchmark the performance of our method on the social media dataset. Additionally we added \textbf{GP+Social} which extends a GP model by adding features extracted from social media data of users, concatenating all features to form a single feature vector. Table~\ref{tbl::performance-results} depicts the result of blood glucose level prediction for different approaches in terms of precision and recall metrics. From the Table, the following points can be observed. (1) As in the previous experiment on CGM data, GP achieves the lowest performance in the BG prediction task. This may be attributed to the fact that it only uses the historical values of BG trajectories to estimate the next BG level. Thus the extreme sparsity of reported values in social media makes it difficult to learn the temporal patterns in BG values of patients (see next section for a detailed study). (2) Although conventional machine learning, i.e., LR and KCCA, outperform the GP approach, they still have low performance according to both metrics. Also as in the previous experiment, the feature-based predictors would overlook temporal dependencies between BG values. Further, KCCA outperforms the LR approach. This can be understood as KCCA performs joint feature selection on both views and builds a latent space from both BG values and social media features. This is in contrast to LR which utilizes early fusion of features, i.e., concatenation of features from both views to form a feature vector. However, KCCA fails to perform optimally as it cannot consider temporal dependencies from both the BG and context data. GP+Social, which uses social media data for prediction, obtains $10\%$ higher performance in both metrics compared to the `conventional' machine learning methods. This is an interesting result and consistent with retrospective studies (\cite{dao2017latent}); demonstrating that social media data of users implicitly indicates health status of patients (e.g. there is a latent representation of health based on this data). Besides, leveraging Gaussian process with context achieved higher performance as compared to KCCA, which shows there also exist temporal patterns in posting behaviour/data of users which can effectively improve prediction. GP+Context obtains the highest performance as it can effectively incorporate different views to learn better latent representation as well as capturing temporal dependencies in the blood glucose prediction task.

\subsection{On the Effect of Sparsity}
\label{sec::data-sparsity-study}
While we demonstrated the approach on social media data due to availability of both blood glucose and context information from this source, and potentially different types of context than in the CGM data, the data is still fairly sparse. Accordingly, to investigate the effect of data sparsity on this approach we generated two less sparse datasets. 

To form each of these datasets, we intuitively select users that have more frequent posting behavior; i.e., from the initial cohort of users, we select users who have at least $25$ BG values to form Dataset-25 and those who have $50$ BG values to form Dataset-50. For each user, we used a random one-week window as a hold out set for evaluation and trained the model on the rest of the dataset. Table~\ref{tbl::sparsity-effect} depicts the performance of our model with different sparsity levels. As can be seen from the table, sparsity has an adverse affect on average performance of the model in terms of both precision and recall metrics. Precision may be highest on the overall dataset again due to the low number of such events. However as expected, denser data significantly improves the performance on identifying adverse events of euglycemic, hyperglycemic and overall blood glucose levels.  

\begin{table*}[h]
\centering
\caption{The effect of sparsity on prediction performance.}
\label{tbl::sparsity-effect}
\begin{tabular}{c|c|c|c|c|c|c|c|c|}
\cline{2-9}
                           & \multicolumn{2}{c|}{\textbf{Hypoglycemic}} & \multicolumn{2}{c|}{\textbf{Euglycemic}} & \multicolumn{2}{c|}{\textbf{Hyperglycemic}} & \multicolumn{2}{c|}{\textbf{Overall}} \\ \cline{2-9} 
                           & Pre         & Rec         & Pre         & Rec         & Pre          & Rec         & Pre           & Rec          \\  \hline
\multicolumn{1}{|l|}{\textbf{Data-All}} & $\mathbf{0.86}$ & $0.53$ & $0.88$ & $0.84$ &  $0.72$ & $0.88$  & $0.82$ & $0.80$  \\ \hline
\multicolumn{1}{|l|}{\textbf{Data-25}} & $0.85$ & $0.75$ & $0.81$ & $\mathbf{0.97}$ &  $0.89$ & $0.75$  & $0.85$ & $0.85$  \\ \hline
\multicolumn{1}{|l|}{\textbf{Data-50}}   & $0.85$ & $\mathbf{0.79}$ & $\mathbf{0.92}$ & $0.94$ &  $\mathbf{0.90}$ & $\mathbf{0.89}$  & $\mathbf{0.90}$ & $\mathbf{0.90}$  \\ \hline
\end{tabular}
\end{table*}

\subsection{Importance of Contextual Information}

We also investigate the effectiveness of different contextual signals in forecasting future blood glucose values. This is important for informing future studies of, and accounting for causal mechanisms of contextual information in relation to blood glucose values. To accomplish this for the CGM data we used a forward step-wise procedure, to decrease computation time. We utilized historical BG values as a base information signal and add one signal each time as contextual signal. Performance improvement obtained by adding each contextual information in regression task is reported in Figure~\ref{fig::signal-study} (Appendix A). From the Figure, we have the following observations: (1) Adding contextual information to BG values decreases the prediction error with a large margin (11.6\% lower RMS error as compared to no contextual information). (2) Different contextual information has varying effects on the RMS error with ``Meal'' and ``Exercise'' having the highest impact on performance (understandably, given their more direct relationship with glucose). Overall, the results demonstrate that incorporating a variety of contextual information can improve the prediction performance and further studies are needed to find the causal mechanisms between these information and blood glucose. For the CGM data, the top ranked features were: \textit{$BG_1, BG_2, BG_5, BG_3, basal, carbs, hypocorrection, snack, BG_6, BG_4$} where $BG_1$, $BG_2$, etc. are the values of blood glucose in historical windows of the previous 30 minutes, $BG_1$ being the most recent one. ``Carbs'' is the self-reported value which shows amount of carbohydrates and ``basal'' is a value that shows amount of insulin intake for correction of blood glucose just before meals. ``Hypocorrection'' is a meal that a user takes for correction of hypoglycemia. 

We also examine the top effective features from social media in predicting future values of BG. We first normalized the weight vector of a linear model for predicting blood glucose values and then ranked the top $10$ features in the trained model. The top ranked features were: \textit{ ``ha'', ``low'', ``lovely'', ``worse'', ``proud'', ``bad'', ``bolus'', ``gym'', 'cheese', ``coffee''}. It should be noted that all of the top features came from unigram text features. Moreover, these top social signal can be categorized in three different categories: emotions (ha, lovely), lifestyle (cheese, coffee, gym), and medical information (bolus). Worse, low and bad may be related to insulin directly or other emotions.

\section{Conclusion}
This paper introduced a method for using, for the first time, contextual information in tandem with blood glucose values, for prediction of high/low/normal blood glucose events (social media data), as well as continuous blood glucose values (CGM data). This work differs from existing work on blood glucose prediction as we focus on the challenges of 1) integrating contextual data as multiple views of information (to add context to the blood glucose value prediction problem) and 2) examine the performance of different kinds of contextual data, from those that provide both multi-view, and multi-modal variables that are not as consistent or specific as blood glucose data. Especially in the case of social media features, the context data can be sparse and noisy. To address all of these challenges, we leveraged Gaussian Processes for understanding the latent temporal representations and prediction in this novel multi-signal task. 

We conclude that contextual information can improve blood glucose prediction in addition to just glucose value data, and those features directly known to relate to blood glucose. Examining the importance of features also provides relevant findings regarding social context and glucose concentrations that can be explored and/or harnessed further. For example, in the CGM data, the self-reported amount of carbohydrates (``carbs'') and amount of insulin intake for correction of blood glucose just before meals (``basal'') are important features. While those two items are clearly directly related to blood glucose, also self-reporting of eating a meal that may be taken specifically in cases of hypoglycemia (``hypocorrection'') is also more informative than some previous blood glucose values themselves. Other less commonly examined features including self-reported sleep, work, and physiological measures such as heart rate, GSR and skin temperature can also improve blood glucose prediction, and indicates that studies of the causal mechanisms should be pursued. We have not interpreted the relative positioning of the different blood glucose windows (i.e. why the fifth previous window is more informative than the fourth), however cyclical behavior in blood glucose may be related and should be studied further. In the social media data, the top features all related to text. while some have a clear link to insulin, diet or activity (e.g. ``bolus'', ``gym'', ``cheese'', ``coffee''), it is worth noting the specificity of some of the diet items. Moreover, several of the top six features are all more related to sentiment or mood (``ha'', ``lovely'', ``proud''). As far as we can tell, this is the first empirical study of such mood features in relation to blood glucose, and this also warrants further systematic study on these features specifically.

While applied to the problem of blood glucose prediction, this approach could be applied to other medical time series prediction problems where multiple views, such as from social context, are available and relevant. Results show that incorporating contextual information can improve blood glucose value prediction above just the glucose values (in sparse data), and where less sparse data is available this could still add some value. Accordingly, in future work, it would be very useful to devise an approach to garner continuous blood glucose information alongside more continuous contextual/social information; perhaps by designing a new type of CGM. Then, more continuous values for specific features can also be studied for their mechanistic relationships with blood glucose. The work could also be used to combine low-level (e.g. physiologic measures) and high-level (social, contextual information) collected by sensors such as wearable and quantified-self devices, to forecast future health states of individuals in a comprehensive manner. Indeed, as this method exploits Gaussian processes, a well-known non-parametric nonlinear approach for time series forecasting, for learning the latent space from input variables, it can effectively model time series data collected by wearables and other devices which are used in daily life and can be linked and combined with contextual information to improve predictions as well. 

In a continuous forecasting scenario of blood glucose level prediction, multi-step prediction is often required, and incorporating prior information of periodicities can be performed by utilizing appropriate kernel functions. While we utilized the well-known Radial Basis Function (RBF) kernel due to its proven effectiveness in several applications, such a simple covariance function may not fully capture the unique characteristics of blood glucose variations in all real-life scenarios. Thus combinations of different kernel functions could be explored to model a number of effects together (\cite{karunaratne2017modelling}). For example, a squared exponential kernel coupled with a periodic kernel can induce a recency effect to the periodic kernel. While we leave this direction for future research, it could potentially  improve the performance of blood glucose level prediction given the multiple types of contexts and their temporal periodicities that can affect glucose values.

%% file: appendix.tex
\section*{Acknowledgments}
This work was partially supported by grant IIS-1845487 from the National Science Foundation.
\section*{Appendix A}
\subsection*{Model Performance}

\begin{figure}[h]
\centering
\includegraphics[width=0.8\columnwidth]{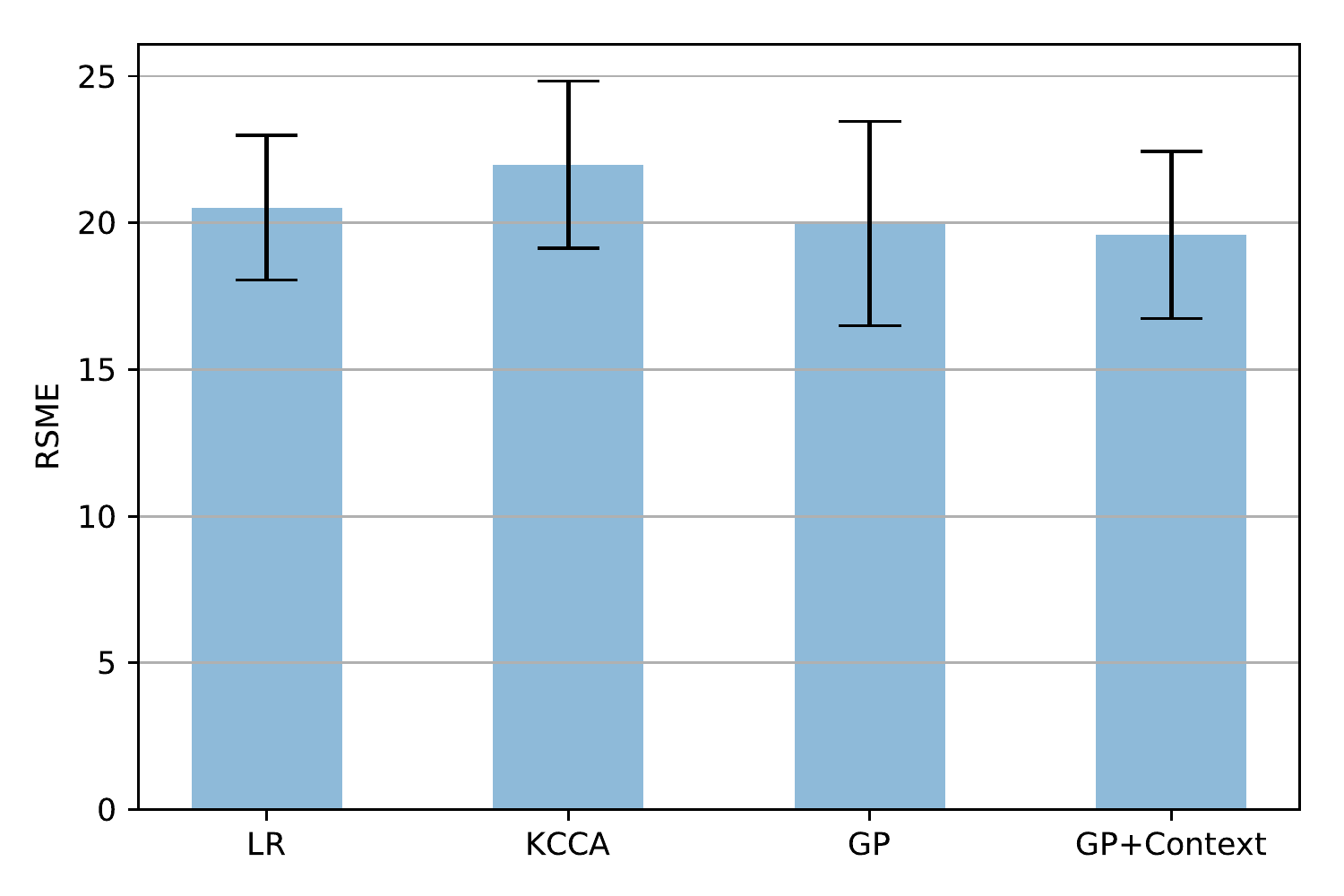}
\caption{Comparison of different methods in prediction future blood glucose values on the CGM dataset.}
\label{fig::performancre-result-cgm-regression}
\end{figure}

\begin{figure}[h]
\centering
\includegraphics[width=0.6\columnwidth]{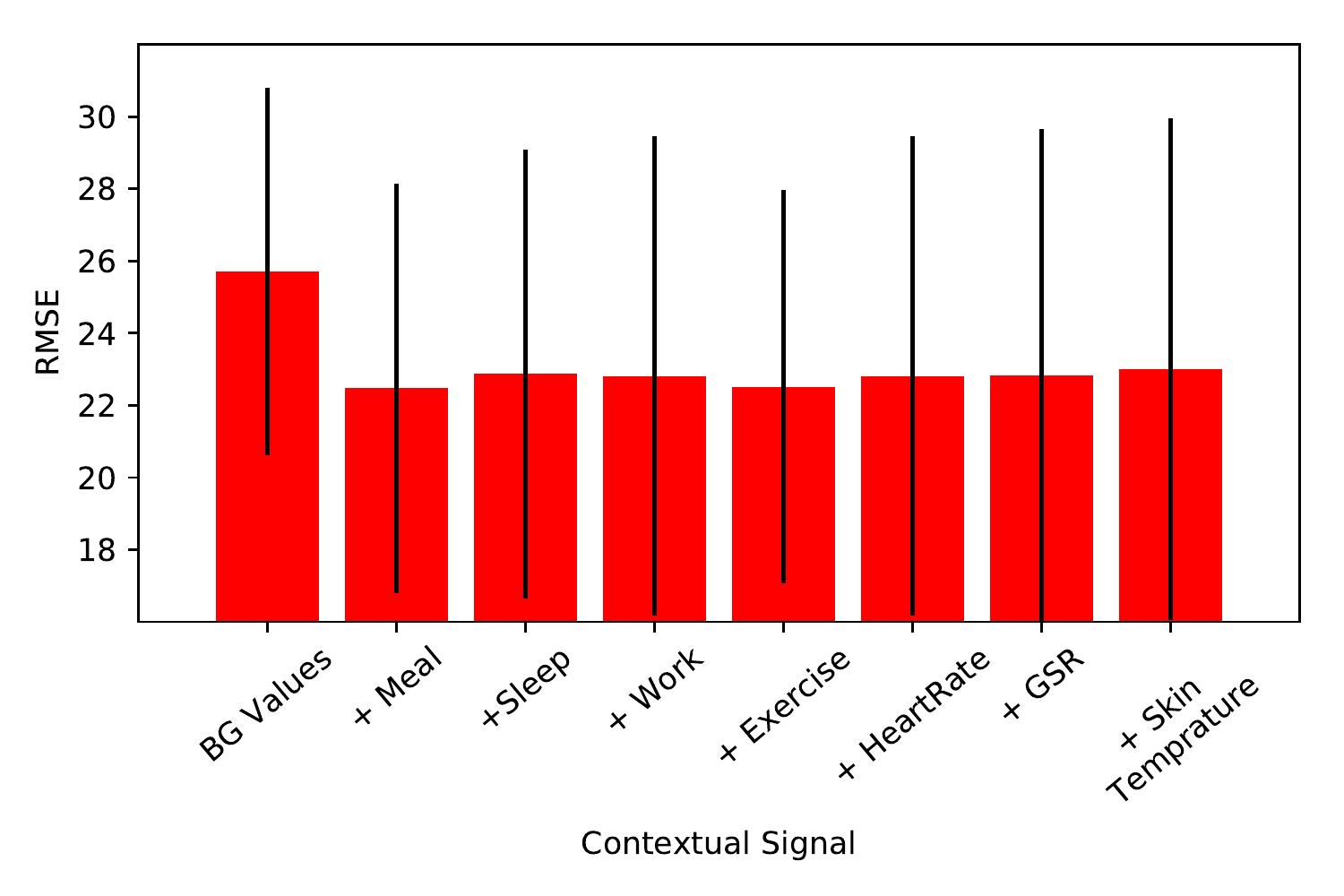}
\caption{The effects of distinct contextual information in CGM data}
\label{fig::signal-study}
\end{figure}

\section*{Appendix B}
\subsection*{Feature Representation}


The contextual information available in social media platforms are available in form of unstructured text, i.e., users post on the network. We extract social context of users from these posts. To represent user context we extracted two types of features: user-centric and content-centric features. 

\paratitle{User-centric Features.}
User-centeric features represent a user's behavior and characteristics. These features are directly extracted from user timelines and demonstrate a users' online posting behaviours. For instance, this would include features such as the number of posted tweets or user-mentions (of a user in the tweets of others via @user-name). User-centric features can implicitly represent user characteristics or differences at specific times (e.g. how active they are, etc.). Table~\ref{tbl::non-text-attributes} presents six distinct attributes we extracted and used in our prediction task.

\begin{table}[ht]
\centering
\caption{User-centric features extracted from timelines.}
\begin{tabular}{|c|l|}
\hline
Feature & Description \\ \hline
$u_1$   &   proportion of retweets          \\ \hline
$u_2$   &   avg. number of retweets/tweet          \\ \hline
$u_3$   &   proportion of hashtags          \\ \hline
$u_4$   &   proportion of tweets with hashtags          \\ \hline
$u_5$   &   avg. number of tweets/day          \\ \hline
$u_6$   &   total number of tweets          \\ \hline
\end{tabular}
\label{tbl::non-text-attributes}
\end{table}

\paratitle{Content-centric Features.}
Before extracting content-centeric features, we conducted a series of common prepossessing steps in order to reduce noise and sparsity in data. We first converted all words to lower case form, removed stop words, replaced emoticons with words and performed stemming. We replaced multiple occurrences of a character with a double occurrence, to correct for misspellings and lengthening, e.g., looool (\cite{agarwal2011sentiment}). We also removed simple re-tweets from the training set in order to remove potential bias (\cite{llewellyn2014re}). Different content-based features are proposed for representing social texts. Here we use the commonly-used  Uni-gram as a base textual feature. In addition, inspired by recent advancements in word embeddings (\cite{mikolov2013efficient,brown1992class}),  we extracted two other text-based features which can better represent the topics discussed by users.

\paratitle{1. Unigrams.} Content-centric features are employed to extract implicit states and behaviors of users from their posts in social media platforms. We thus process the tweets published by users and represent each user based on these features. As a basic feature we used uni-grams for representations. To do so, we first form a a vocabulary of size $\vert V \vert = 12,904$ where we only retain the terms which appear more than $10$ times in the corpus. All term frequencies are normalized with the total number of tweets posted by the user. 




\paratitle{2. Neural Embeddings (W2V-E).}
Recently, there has been a growing interest in neural language models, wherein words are projected into a lower dimensional dense vector space via a hidden layer (\cite{mikolov2013efficient,mikolov2013distributed}). These models can provide a better representation of words compared to traditional language models because they can also capture syntactic information rather than just bag-of-words, handling non-linear transformations (\cite{mikolov2013efficient}). In this low dimensional vector space, words with a small distance are considered semantically similar. We use the skip-gram model with negative sampling to learn word embeddings on the Twitter reference corpus (\cite{mikolov2013efficient}). We use a layer size of 50 and the Gensim implementation\footnote{\url{https://radimrehurek.com/gensim/}}.



\paratitle{3. Word Brown Clusters}
Word clustering techniques, such as Brown Clustering~\cite{brown1992class}, have proven to be effective in mitigating term sparsity in open-domain and domain-specific tasks~\cite{ratinov2009design,yao2016mobile,ye2016software}. This approach assumes that similar words should appear in similar contexts. At each iteration, it merges semantically similar words into a fixed number of classes based on the log-probability and incurs the least loss in global mutual information.

%% file: main.bbl
\begin{thebibliography}{34}
\providecommand{\natexlab}[1]{#1}
\providecommand{\url}[1]{\texttt{#1}}
\expandafter\ifx\csname urlstyle\endcsname\relax
  \providecommand{\doi}[1]{doi: #1}\else
  \providecommand{\doi}{doi: \begingroup \urlstyle{rm}\Url}\fi

\bibitem[Agarwal et~al.(2011)Agarwal, Xie, Vovsha, Rambow, and
  Passonneau]{agarwal2011sentiment}
Apoorv Agarwal, Boyi Xie, Ilia Vovsha, Owen Rambow, and Rebecca Passonneau.
\newblock Sentiment analysis of twitter data.
\newblock In \emph{Proceedings of the workshop on languages in social media},
  pages 30--38. Association for Computational Linguistics, 2011.

\bibitem[Akaho(2006)]{akaho2006kernel}
Shotaro Akaho.
\newblock A kernel method for canonical correlation analysis.
\newblock \emph{arXiv preprint cs/0609071}, 2006.

\bibitem[Akbari et~al.(2018)Akbari, Relia, Elghafari, and
  Chunara]{akbari2018user}
Mohammad Akbari, Kunal Relia, Anas Elghafari, and Rumi Chunara.
\newblock From the user to the medium: Neural profiling across web communities.
\newblock In \emph{Twelfth International AAAI Conference on Web and Social
  Media}, 2018.

\bibitem[Albers et~al.(2017)Albers, Levine, Gluckman, Ginsberg, Hripcsak, and
  Mamykina]{albers2017personalized}
David~J Albers, Matthew Levine, Bruce Gluckman, Henry Ginsberg, George
  Hripcsak, and Lena Mamykina.
\newblock Personalized glucose forecasting for type 2 diabetes using data
  assimilation.
\newblock \emph{PLoS computational biology}, 13\penalty0 (4):\penalty0
  e1005232, 2017.

\bibitem[Brown et~al.(1992)Brown, Desouza, Mercer, Pietra, and
  Lai]{brown1992class}
Peter~F Brown, Peter~V Desouza, Robert~L Mercer, Vincent J~Della Pietra, and
  Jenifer~C Lai.
\newblock Class-based n-gram models of natural language.
\newblock \emph{Computational linguistics}, 18\penalty0 (4):\penalty0 467--479,
  1992.

\bibitem[Chung et~al.(2018)Chung, Kim, Lee, Hwang, and Yang]{chung2018mixed}
Ingyo Chung, Saehoon Kim, Juho Lee, Sung~Ju Hwang, and Eunho Yang.
\newblock Mixed effect composite rnn-gp: A personalized and reliable prediction
  model for healthcare.
\newblock \emph{arXiv preprint arXiv:1806.01551}, 2018.

\bibitem[Dao et~al.(2017)Dao, Nguyen, Venkatesh, and Phung]{dao2017latent}
Bo~Dao, Thin Nguyen, Svetha Venkatesh, and Dinh Phung.
\newblock Latent sentiment topic modelling and nonparametric discovery of
  online mental health-related communities.
\newblock \emph{International Journal of Data Science and Analytics},
  4\penalty0 (3):\penalty0 209--231, 2017.

\bibitem[De~Choudhury et~al.(2014)De~Choudhury, Morris, and
  White]{de2014seeking}
Munmun De~Choudhury, Meredith~Ringel Morris, and Ryen~W White.
\newblock Seeking and sharing health information online: comparing search
  engines and social media.
\newblock In \emph{Proceedings of the SIGCHI conference on human factors in
  computing systems}, pages 1365--1376. ACM, 2014.

\bibitem[Fox et~al.(2018)Fox, Ang, Jaiswal, Pop-Busui, and Wiens]{fox2018deep}
Ian Fox, Lynn Ang, Mamta Jaiswal, Rodica Pop-Busui, and Jenna Wiens.
\newblock Deep multi-output forecasting: Learning to accurately predict blood
  glucose trajectories.
\newblock \emph{arXiv preprint arXiv:1806.05357}, 2018.

\bibitem[Futoma et~al.(2017)Futoma, Hariharan, and Heller]{futoma2017learning}
Joseph Futoma, Sanjay Hariharan, and Katherine Heller.
\newblock Learning to detect sepsis with a multitask gaussian process rnn
  classifier.
\newblock \emph{arXiv preprint arXiv:1706.04152}, 2017.

\bibitem[Huang et~al.(2017)Huang, Elghafari, Relia, and Chunara]{huang2017high}
Tom Huang, Anas Elghafari, Kunal Relia, and Rumi Chunara.
\newblock High-resolution temporal representations of alcohol and tobacco
  behaviors from social media data.
\newblock \emph{Proceedings of the ACM on human-computer interaction},
  1\penalty0 (CSCW):\penalty0 54, 2017.

\bibitem[Kapoor and Picard(2005)]{kapoor2005multimodal}
Ashish Kapoor and Rosalind~W Picard.
\newblock Multimodal affect recognition in learning environments.
\newblock In \emph{Proceedings of the 13th annual ACM international conference
  on Multimedia}, pages 677--682. ACM, 2005.

\bibitem[Karunaratne et~al.(2017)Karunaratne, Moshtaghi, Karunasekera, Harwood,
  and Cohn]{karunaratne2017modelling}
Pasan Karunaratne, Masud Moshtaghi, Shanika Karunasekera, Aaron Harwood, and
  Trevor Cohn.
\newblock Modelling the working week for multi-step forecasting using gaussian
  process regression.
\newblock In \emph{Proceedings of the 26th International Joint Conference on
  Artificial Intelligence}, pages 1994--2000. AAAI Press, 2017.

\bibitem[Llewellyn et~al.(2014)Llewellyn, Grover, Oberlander, and
  Klein]{llewellyn2014re}
Clare Llewellyn, Claire Grover, Jon Oberlander, and Ewan Klein.
\newblock Re-using an argument corpus to aid in the curation of social media
  collections.
\newblock In \emph{LREC}, volume~14, pages 462--468, 2014.

\bibitem[Lustman and Clouse(2005)]{lustman2005depression}
Patrick~J Lustman and Ray~E Clouse.
\newblock Depression in diabetic patients: the relationship between mood and
  glycemic control.
\newblock \emph{Journal of Diabetes and its Complications}, 19\penalty0
  (2):\penalty0 113--122, 2005.

\bibitem[Marling and Bunescu(2018)]{marling2018ohiot1dm}
Cindy Marling and Razvan Bunescu.
\newblock The ohiot1dm dataset for blood glucose level prediction.
\newblock In \emph{The 3rd International Workshop on Knowledge Discovery in
  Healthcare Data, Stockholm, Sweden}, 2018.

\bibitem[Mignault et~al.(2005)Mignault, Onge, Karelis, Allison, and
  Rabasa-Lhoret]{mignault2005evaluation}
Diane Mignault, Maxime~St Onge, Antony~D Karelis, David~B Allison, and Remi
  Rabasa-Lhoret.
\newblock Evaluation of the portable healthwear armband: a device to measure
  total daily energy expenditure in free-living type 2 diabetic individuals.
\newblock \emph{Diabetes care}, 28\penalty0 (1):\penalty0 225--227, 2005.

\bibitem[Mikolov et~al.(2013{\natexlab{a}})Mikolov, Chen, Corrado, and
  Dean]{mikolov2013efficient}
Tomas Mikolov, Kai Chen, Greg Corrado, and Jeffrey Dean.
\newblock Efficient estimation of word representations in vector space.
\newblock \emph{arXiv preprint arXiv:1301.3781}, 2013{\natexlab{a}}.

\bibitem[Mikolov et~al.(2013{\natexlab{b}})Mikolov, Sutskever, Chen, Corrado,
  and Dean]{mikolov2013distributed}
Tomas Mikolov, Ilya Sutskever, Kai Chen, Greg~S Corrado, and Jeff Dean.
\newblock Distributed representations of words and phrases and their
  compositionality.
\newblock In \emph{Advances in neural information processing systems}, pages
  3111--3119, 2013{\natexlab{b}}.

\bibitem[Oviedo et~al.(2017)Oviedo, Veh{\'\i}, Calm, and
  Armengol]{oviedo2017review}
Silvia Oviedo, Josep Veh{\'\i}, Remei Calm, and Joaquim Armengol.
\newblock A review of personalized blood glucose prediction strategies for t1dm
  patients.
\newblock \emph{International journal for numerical methods in biomedical
  engineering}, 33\penalty0 (6):\penalty0 e2833, 2017.

\bibitem[Plis et~al.(2014)Plis, Bunescu, Marling, Shubrook, and
  Schwartz]{plis2014machine}
Kevin Plis, Razvan~C Bunescu, Cindy Marling, Jay Shubrook, and Frank Schwartz.
\newblock A machine learning approach to predicting blood glucose levels for
  diabetes management.
\newblock In \emph{AAAI Workshop: Modern Artificial Intelligence for Health
  Analytics}, number~31, pages 35--39, 2014.

\bibitem[Rasmussen(2004)]{rasmussen2004gaussian}
Carl~Edward Rasmussen.
\newblock Gaussian processes in machine learning.
\newblock In \emph{Advanced lectures on machine learning}, pages 63--71.
  Springer, 2004.

\bibitem[Ratinov and Roth(2009)]{ratinov2009design}
Lev Ratinov and Dan Roth.
\newblock Design challenges and misconceptions in named entity recognition.
\newblock In \emph{Proceedings of the Thirteenth Conference on Computational
  Natural Language Learning}, pages 147--155. Association for Computational
  Linguistics, 2009.

\bibitem[Sato et~al.(2003)Sato, Nagasaki, Nakai, and Fushimi]{sato2003physical}
Yuzo Sato, Masaru Nagasaki, Naoya Nakai, and Takashi Fushimi.
\newblock Physical exercise improves glucose metabolism in lifestyle-related
  diseases.
\newblock \emph{Experimental Biology and Medicine}, 228\penalty0 (10):\penalty0
  1208--1212, 2003.

\bibitem[Shumway and Stoffer(2011)]{shumway2011time}
Robert~H Shumway and David~S Stoffer.
\newblock Time series regression and exploratory data analysis.
\newblock In \emph{Time series analysis and its applications}, pages 47--82.
  Springer, 2011.

\bibitem[Snoek et~al.(2005)Snoek, Worring, and Smeulders]{snoek2005early}
Cees~GM Snoek, Marcel Worring, and Arnold~WM Smeulders.
\newblock Early versus late fusion in semantic video analysis.
\newblock In \emph{Proceedings of the 13th annual ACM international conference
  on Multimedia}, pages 399--402. ACM, 2005.

\bibitem[Sparacino et~al.(2007)Sparacino, Zanderigo, Corazza, Maran,
  Facchinetti, and Cobelli]{sparacino2007glucose}
Giovanni Sparacino, Francesca Zanderigo, Stefano Corazza, Alberto Maran, Andrea
  Facchinetti, and Claudio Cobelli.
\newblock Glucose concentration can be predicted ahead in time from continuous
  glucose monitoring sensor time-series.
\newblock \emph{IEEE Transactions on biomedical engineering}, 54\penalty0
  (5):\penalty0 931--937, 2007.

\bibitem[Tang et~al.(2013)Tang, Hu, Gao, and Liu]{tang2013unsupervised}
Jiliang Tang, Xia Hu, Huiji Gao, and Huan Liu.
\newblock Unsupervised feature selection for multi-view data in social media.
\newblock In \emph{Proceedings of the 2013 SIAM International Conference on
  Data Mining}, pages 270--278. SIAM, 2013.

\bibitem[Thompson(2005)]{thompson2005canonical}
Bruce Thompson.
\newblock Canonical correlation analysis.
\newblock \emph{Encyclopedia of statistics in behavioral science}, 2005.

\bibitem[Valletta et~al.(2009)Valletta, Chipperfield, and
  Byrne]{valletta2009gaussian}
John~Joseph Valletta, Andrew~J Chipperfield, and Christopher~D Byrne.
\newblock Gaussian process modelling of blood glucose response to free-living
  physical activity data in people with type 1 diabetes.
\newblock In \emph{2009 Annual International Conference of the IEEE Engineering
  in Medicine and Biology Society}, pages 4913--4916. IEEE, 2009.

\bibitem[Xu et~al.(2011)Xu, Bu, Chen, Cai, He, Liu, and Luo]{xu2011efficient}
Bin Xu, Jiajun Bu, Chun Chen, Deng Cai, Xiaofei He, Wei Liu, and Jiebo Luo.
\newblock Efficient manifold ranking for image retrieval.
\newblock In \emph{Proceedings of the 34th international ACM SIGIR conference
  on Research and development in Information Retrieval}, pages 525--534. ACM,
  2011.

\bibitem[Yao and Sun(2016)]{yao2016mobile}
Yangjie Yao and Aixin Sun.
\newblock Mobile phone name extraction from internet forums: a semi-supervised
  approach.
\newblock \emph{World Wide Web}, 19\penalty0 (5):\penalty0 783--805, 2016.

\bibitem[Ye et~al.(2016)Ye, Xing, Foo, Ang, Li, and Kapre]{ye2016software}
Deheng Ye, Zhenchang Xing, Chee~Yong Foo, Zi~Qun Ang, Jing Li, and Nachiket
  Kapre.
\newblock Software-specific named entity recognition in software engineering
  social content.
\newblock In \emph{Software Analysis, Evolution, and Reengineering (SANER),
  2016 IEEE 23rd International Conference on}, volume~1, pages 90--101. IEEE,
  2016.

\bibitem[Ye et~al.(2012)Ye, Liu, Jhuo, and Chang]{ye2012robust}
Guangnan Ye, Dong Liu, I-Hong Jhuo, and Shih-Fu Chang.
\newblock Robust late fusion with rank minimization.
\newblock In \emph{Computer Vision and Pattern Recognition (CVPR), 2012 IEEE
  Conference on}, pages 3021--3028. IEEE, 2012.

\end{thebibliography}
